\documentclass{article}
\usepackage{spconf,amsmath,graphicx}
\usepackage{hyperref}
\usepackage{amssymb}
\usepackage{amsmath}
\usepackage{multirow}
\usepackage{booktabs}
\usepackage{xcolor}
\usepackage{pifont}

\title{WMoE-CLIP: Wavelet-Enhanced Mixture-of-Experts Prompt Learning for Zero-Shot Anomaly Detection}
\name{Peng Chen, Chao Huang*\thanks{This work was supported in part by the National Natural Science Foundation of China (No.62301621), Shenzhen Science and Technology Program (No. 20231121172359002, 2023A008), Shenzhen General Research Project (No. JCYJ20241202125904007), and Guangdong Basic and Applied Basic Research Foundation (No. 2025A1515011398).}}
\address{School of Cyber Science and Technology, Shenzhen Campus of Sun Yat-sen University, Shenzhen\\
23pchen@stu.edu.cn, huangch253@mail.sysu.edu.cn}
\begin{document}
%
\maketitle
\begin{abstract}
Vision–language models have recently shown strong generalization in zero-shot anomaly detection (ZSAD), enabling the detection of unseen anomalies without task-specific supervision. However, existing approaches typically rely on fixed textual prompts, which struggle to capture complex semantics, and focus solely on spatial-domain features, limiting their ability to detect subtle anomalies. To address these challenges, we propose a wavelet-enhanced mixture-of-experts prompt learning method for ZSAD. Specifically, a variational autoencoder is employed to model global semantic representations and integrate them into prompts to enhance adaptability to diverse anomaly patterns. Wavelet decomposition extracts multi-frequency image features that dynamically refine textual embeddings through cross-modal interactions. Furthermore, a semantic-aware mixture-of-experts module is introduced to aggregate contextual information. Extensive experiments on 14 industrial and medical datasets demonstrate the effectiveness of the proposed method.
\end{abstract}
\section{Introduction}

Anomaly detection aims to identify instances that deviate from normal patterns and has been widely applied across various domains, including industrial~\cite{jeong2023winclip,ma2025aa} and medical scenarios~\cite{jha2019kvasir,cao2024adaclip}. Due to the diversity of anomaly types and the scarcity of abnormal samples, traditional approaches typically adopt unsupervised strategies. However, in practical applications, acquiring sufficient training data remains a major challenge~\cite{chen2026reason}. For example, patient data may be restricted by privacy regulations~\cite{zhou2023anomalyclip}, and new production lines may lack representative samples. ZSAD addresses this by detecting unseen anomalies using auxiliary datasets. However, achieving robust generalization across diverse product categories and complex environments remains a persistent challenge.

Vision-language models like CLIP have advanced ZSAD by leveraging large-scale image–text pretraining~\cite{ma2025aa}, enabling robust generalization to unseen categories~\cite{chen2026dyc}. For example, WinCLIP~\cite{jeong2023winclip} identifies potential anomalous regions by measuring the similarity between multi-scale visual features and manually crafted textual prompts. To reduce reliance on handcrafted prompts, AnomalyCLIP~\cite{zhou2023anomalyclip} introduces learnable class-agnostic textual prompts with a V-V attention, enhancing zero-shot performance. Furthermore, AdaCLIP~\cite{cao2024adaclip} employs a hybrid prompting strategy that combines dynamic and static prompts to improve generalization, while AA-CLIP~\cite{ma2025aa} leverages a two-stage training paradigm to better adapt CLIP for anomaly detection. However, two critical issues remain. First, fixed prompts provide sparse semantic information, potentially leading to overfitting within a constrained semantic space. Second, exclusive reliance on spatial information limits the model’s ability to detect subtle defects.

To address these challenges, we propose WMoE-CLIP, a novel wavelet-enhanced mixture-of-experts prompt learning approach based on CLIP for ZSAD. Our method improves anomaly detection by reinforcing the alignment and adaptability of image–text representations. Specifically, a variational autoencoder (VAE) is employed to sample from the global feature distribution, embedding rich semantic information into the prompts. To further strengthen multimodal alignment, wavelet decomposition extracts multi-frequency image features, which dynamically refine textual embeddings via cross-attention, enabling the detection of subtle anomalies. Additionally, a semantic-aware mixture-of-experts module aggregates contextual information, further enhancing the model’s capability to recognize diverse anomaly patterns.

In summary, our contributions are as follows: 1) We propose WMoE-CLIP, a novel CLIP-based method that enhances image–text interactions, effectively improving both the accuracy and generalization of zero-shot anomaly detection. 2) We leverage a VAE to model global feature distributions and incorporate frequency-domain features to strengthen cross-modal interactions. Additionally, we introduce a semantic-aware mixture-of-experts module to aggregate contextual semantic information. 3) Extensive experiments on 14 industrial and medical datasets demonstrate that WMoE-CLIP achieves state-of-the-art performance.

\begin{figure*}
    \centering
    \includegraphics[width=0.85\linewidth]{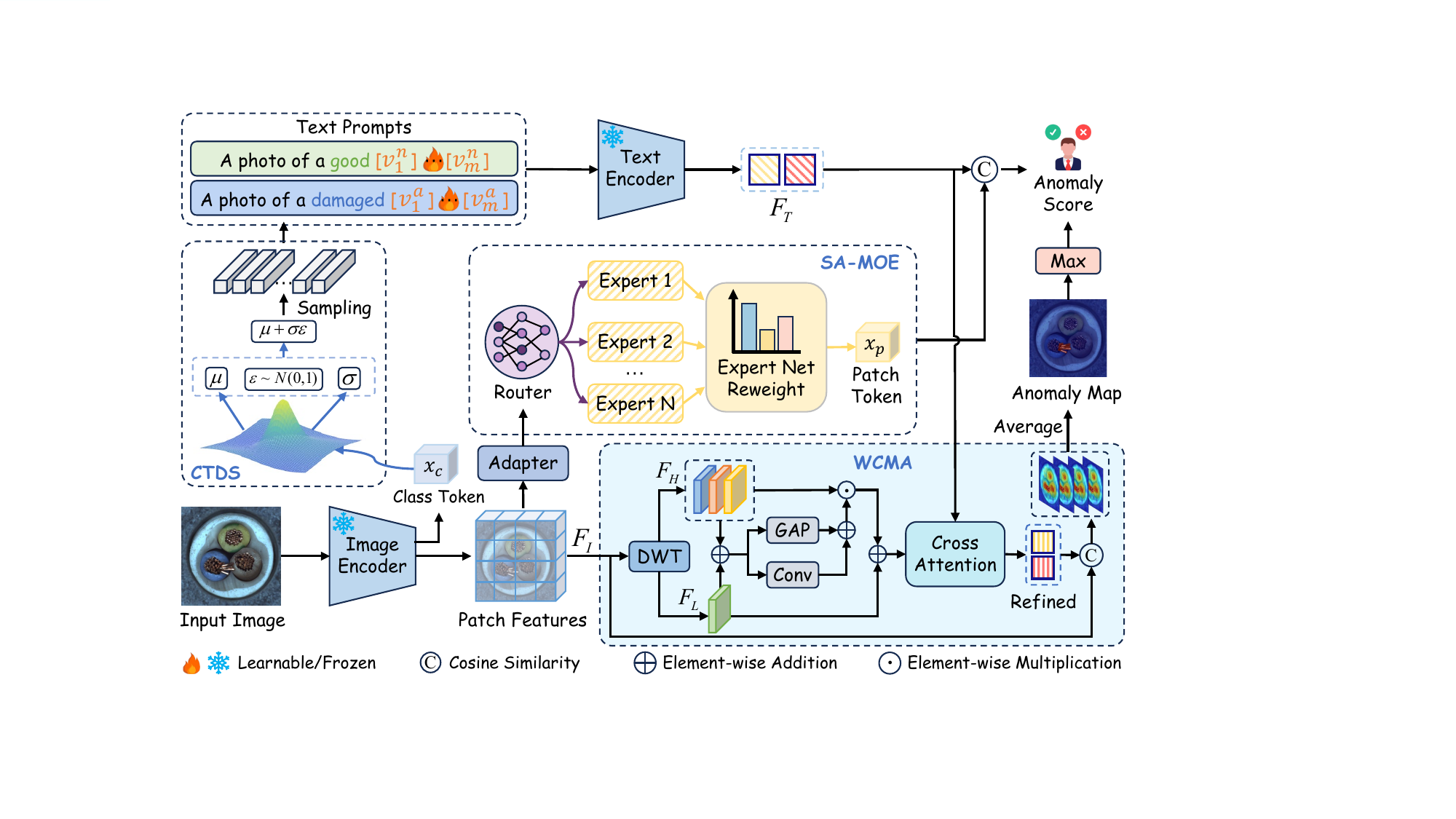}
    \caption{\textbf{Framework of WMoE-CLIP}. CTDS leverages a VAE to model global semantic features, enhancing the adaptability of text embeddings. WCMA dynamically updates text embeddings using wavelet-based frequency features. SA-MoE employs a mixture-of-experts model to capture rich contextual information, further enhancing robust image-level anomaly scoring.}
    \label{fig:method}
\end{figure*}

\section{Methodology}

In this paper, we propose WMoE-CLIP, a wavelet-enhanced mixture-of-experts model built upon CLIP for ZSAD. As illustrated in Fig.~\ref{fig:method}, WMoE-CLIP comprises three core components: Class Token Distribution Sampling (CTDS), Wavelet-Enhanced Cross-Modal Attention (WCMA), and Semantic-Aware Mixture-of-Experts (SA-MoE). Given an input image $I\in\mathbb{R}^{h\times w \times 3}$, a frozen image encoder extracts the global class token $x_c\in\mathbb{R}^{C}$ and patch features $F_I\in\mathbb{R}^{HW \times C}$, where $C$ denotes the embedding dimension. CTDS models the latent distribution of the global features and integrates them into the prompts. To facilitate cross-modal alignment, WCMA dynamically refines text embeddings guided by frequency-domain image features. Furthermore, SA-MoE adaptively selects experts to aggregate contextual semantic information.

\subsection{Class Token Distribution Sampling}
Existing approaches typically rely on coarse-grained prompts, which lack adaptability to image-specific visual contexts. To address this limitation, we propose CTDS, which leverages a VAE to model the global distribution and integrate the sampled representations into learnable prompt embeddings. As shown in Fig.~\ref{fig:method}, the global class token $x_c$ from the image encoder is projected into a latent space via two fully connected layers to obtain the mean and variance, from which a latent variable is sampled via reparameterization, formulated as:
\begin{equation}
    \begin{aligned}
        \mu_c &= W_{\mu}\text{MLP}(x_c),\quad \log \sigma_c ^2=W_{\sigma}\text{MLP}(x_c),\\
        s_c &= \mu_c+\epsilon \odot \sigma_c, \quad \epsilon \sim \mathcal{N}(0, I),
    \end{aligned}
\end{equation}
where $W_{\mu}, W_{\sigma}$ are learnable parameters, and $\odot$ represents element-wise multiplication. The latent representation $s_c$ is passed through a decoder (an MLP) to obtain the reconstructed global feature $r\in\mathbb{R}^{C}$, and the latent distribution is regularized using the Kullback-Leibler divergence:
\begin{align}
    \mathcal{L}_{\text{KL}} = -\frac{1}{2}\sum_{i=1}^{C} \Big( 1 + \log (\sigma_c^i)^2 - (\mu_c^i)^2 - (\sigma_c^i)^2 \Big).
\end{align}

In addition, to encourage consistency between the original and reconstructed features, we employ a reconstruction loss:
\begin{align}
    \mathcal{L}_{\text{rec}} = \| r - x_c \|^2.
\end{align}

We introduce the template ``a photo of a good/damaged $[v_i^n]/[v_i^a]$'', where $[v_i^n]$ and $[v_i^a]$ are learnable vectors encoding category-specific semantics. $\{r_i\}_{i=1}^m$ are sampled from the latent space and are fused with the corresponding vector as $g(r_i,v_i)=r_i+v_i$, which replaces $v_i$ in the prompt. The normal and abnormal prompts, $P^n$ and $P^a$, are defined as:
\begin{equation}
\begin{aligned}
P^n &= \text{a photo of a good } \; g(r_1, v_1^n), \dots, g(r_m, v_m^n), \\
P^a &= \text{a photo of a damaged } \; g(r_1, v_1^a), \dots, g(r_m, v_m^a).
\end{aligned}
\end{equation}

The constructed prompts are fed into the text encoder to obtain the corresponding text embeddings $F_T\in\mathbb{R}^{2 \times C}$.

\subsection{Wavelet-Enhanced Cross-Modal Attention}
Incorporating image-specific features into text embeddings complicates alignment with image features. Additionally, high-frequency components preserve rich details crucial for detecting subtle anomalies. Therefore, we propose WCMA to enhance the adaptability of text embeddings to image features. Specifically, a Haar wavelet transform is applied to decompose $F_I$ into low-frequency information ($F_{L}$) and high-frequency components along the horizontal ($F_{LH}$), vertical ($F_{HL}$), and diagonal ($F_{HH}$) directions. The high-frequency sub-bands are then aggregated ($F_H = F_{LH} + F_{HL} + F_{HH}$) to form an overall high-frequency component.

To capture high-frequency information, we jointly model global and local dependencies across multi-frequency components. $F_L\in\mathbb{R}^{H\times W \times C}$ and $F_H\in\mathbb{R}^{H\times W \times C}$ are combined, with global channel representations obtained via global average pooling (GAP) and local representations via pointwise convolution. These are then fused to generate the attention weights $W_h$ for the high-frequency component:
\begin{equation}
    \begin{aligned}
        W_h = \sigma\!\big(W_1 \cdot \mathrm{ReLU}(\mathrm{GAP}(F_L+F_H)) \\+ W_2 \cdot \mathrm{ReLU}(\delta_p(F_L+F_H)) \big),
    \end{aligned}
\end{equation}
where $\sigma(\cdot)$ denotes the sigmoid function, $W_1$ and $W_2$ are linear layers, and $\delta_p$ indicates the pointwise convolution. The frequency-enhanced image representation $F_p\in\mathbb{R}^{H\times W \times C}$ is obtained by reweighting high-frequency features and aggregating them with the low-frequency features:
\begin{align}
    F_p = F_H \odot W_h + F_L,
\end{align}
where $\odot$ denotes element-wise multiplication. To adapt text embeddings to patch features, we employ a cross-attention, where text queries interact with the image keys and values:
\begin{align}
    F_T' = \mathrm{softmax}\!\left(Q_T K_I^\top / {\sqrt{C}}\right) V_I,
\end{align}
where $F_T'\in\mathbb{R}^{2 \times C}$ are the refined text embeddings. $Q_T = F_T W_q$, $K_I = F_p W_k$, $V_I = F_p W_v$, and $W_q, W_k, W_v$ are learnable parameters. The anomaly map $M\in\mathbb{R}^{h \times w}$ is computed from the similarity between $F_T'$ and the patch features.

\begin{figure}
    \centering
    \includegraphics[width=0.92\linewidth]{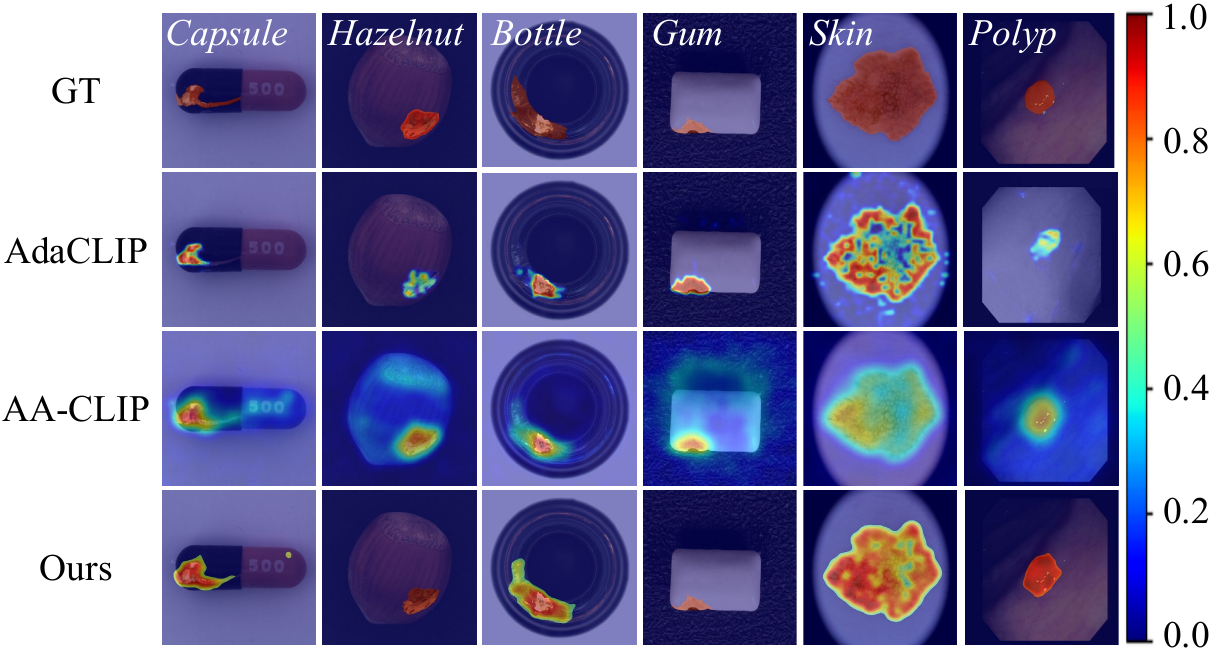}
    \caption{Comparative visualization of anomaly localization. }
    \label{fig:vis}
\end{figure}

\subsection{Semantic-Aware Mixture-of-Experts}
To enable more reliable anomaly score computation, we propose SA-MoE, which adaptively aggregates global semantic information from contextual features. Given patch features from multiple layers, we introduce an Adapter consisting of a linear layer followed by global average pooling. The linear layer projects patch features into a unified space. The features are then concatenated along the channel dimension, and global average pooling across the spatial dimensions yields the contextual representation $x_a\in\mathbb{R}^{C}$.

SA-MoE consists of a routing gate $\mathcal{G}(\cdot)$ and $N$ expert networks $E_n(\cdot)$, which dynamically activate the Top-$k$ experts to aggregate global semantics. Specifically, the routing gate computes a relevance score $s_n = \mathcal{G}(x_a)$ for each expert based on $x_a$. The enhanced patch features are obtained by aggregating the outputs of the selected experts as follows: 
\begin{align}
    x_p=\sum_k w_k E_k(x_a)
\end{align}
where $w_k$ denotes the normalized routing weight of the $k$-th expert. To further consolidate semantic information, $x_p$ is integrated with the class token: $x_{cls}=x_p+x_{c}$. The anomaly score $s_{txt}$ is then computed as the similarity between $x_{cls}$ and the text feature $F_T$. The final image-level anomaly score is obtained by combining $s_{txt}$ with the maximum value of the anomaly map: $\hat{s}=s_{txt}+\text{max}(\text{M})$.

\subsection{Loss Function}
We jointly employ global and local losses to guide optimization. The global loss adopts the binary cross-entropy to supervise image-level predictions, while the local loss combines Focal and Dice losses to optimize pixel-level predictions:
\begin{align}
    \mathcal{L}_{global} &= \mathrm{BCE}(S_{gt}, \hat{s}), \\
    \mathcal{L}_{local} &= \sum_{l} \mathrm{Focal}(M_l, M_{gt}) + \sum_{l} \mathrm{Dice}(M_l, M_{gt}),
\end{align}
where $S_{gt} \in \{0,1\}$ and $M_{gt} \in \mathbb{R}^{h \times w}$ denote the ground-truth label and pixel-level mask. The overall objective function is:
\begin{align}
    \mathcal{L}=\mathcal{L}_{global}+\mathcal{L}_{local}+\mathcal{L}_{KL}+\mathcal{L}_{rec}.
\end{align}

\begin{table*}[]
\caption{\textbf{Comparison of ZSAD performance.} The best results are highlighted in red, and the second-best are marked in blue.}
\label{table_zsad}
\resizebox{\linewidth}{!}{
\begin{tabular}{ccccccccc}
\toprule
\multirow{2}{*}{Domain}      & \multirow{2}{*}{Metric}                                                                     & \multirow{2}{*}{Dataset} & WinCLIP                                 & CLIP-AD                                 & AnomalyCLIP                                                  & Ada-CLIP                                                                          & AA-CLIP                                                                           & WMoE-CLIP                                                                       \\ \cline{4-9} 
                             &                                                                                             &                          & CVPR2023                                & IJCAI2024                               & ICLR2024                                                     & ECCV2024                                                                          & CVPR2025                                                                          & ---                                                                             \\ \hline
\multirow{12}{*}{Industrial} & \multirow{6}{*}{\begin{tabular}[c]{@{}c@{}}Image-level\\ (AUROC,F1-Max,\\ AP)\end{tabular}} & MVTec-AD                 & (91.8, \textcolor{blue}{92.9}, 95.1) & (89.8, 91.1, 95.3)                      & (91.5, 92.8, 96.2)                                           & (\textcolor{blue}{92.0}, 92.7, \textcolor{blue}{96.4})                      & (90.5, 91.4, 95.6)                                                                & (\textcolor{red}{92.4}, \textcolor{red}{93.2}, \textcolor{red}{96.9})  \\
                             &                                                                                             & VisA                     & (78.1, 79.0, 77.5)                      & (79.8, 79.2, 84.3)                      & (82.1, 80.4, \textcolor{blue}{85.4})                      & (83.0, \textcolor{blue}{81.6}, 84.9)                                           & (\textcolor{blue}{84.6}, 78.7, 83.4)                                           & (\textcolor{red}{87.3}, \textcolor{red}{84.4}, \textcolor{red}{90.0})  \\
                             &                                                                                             & BTAD                     & (83.3, 81.0, 84.1)                      & (85.8, 81.7, 85.2)                      & (89.1, 86.0, 91.1)                                           & (91.6, 88.9, 92.4)                                                                & (\textcolor{red}{93.8}, \textcolor{blue}{92.8}, \textcolor{blue}{97.1})  & (\textcolor{blue}{92.6}, \textcolor{red}{95.0}, \textcolor{red}{97.8}) \\
                             &                                                                                             & KSDD2                    & (93.5, 71.4, 77.9)                      & (95.2, 84.4, 88.2)                      & (92.1, 71.0, 77.8)                                           & (\textcolor{blue}{95.9}, 84.5, 95.9)                                           & (95.8, \textcolor{blue}{91.4}, \textcolor{blue}{96.8})                      & (\textcolor{red}{96.8}, \textcolor{red}{91.5}, \textcolor{red}{97.4})  \\
                             &                                                                                             & DAGM                     & (89.6, 86.4, 90.4)                      & (90.8, 88.4, 90.5)                      & (95.6, 93.2, \textcolor{blue}{94.6})                      & (\textcolor{blue}{96.5}, \textcolor{red}{94.1}, 95.7)                       & (94.3, 81.7, 84.2)                                                                & (\textcolor{red}{96.7}, \textcolor{blue}{93.4}, \textcolor{red}{95.9}) \\
                             &                                                                                             & DTD-Synthetic            & (93.2, 94.3, 92.6)                      & (91.5, 91.8, 96.8)                      & (\textcolor{blue}{93.5}, \textcolor{blue}{94.5}, 97.0) & (92.8, 92.2, 97.0)                                                                & (91.4, 93.3, \textcolor{blue}{97.2})                                           & (\textcolor{red}{95.0}, \textcolor{red}{95.6}, \textcolor{red}{98.4})  \\ \cline{2-9} 
                             & \multirow{6}{*}{\begin{tabular}[c]{@{}c@{}}Pixel-level\\ (AUROC,PRO,\\ AP)\end{tabular}}    & MVTec-AD                 & (85.1, 64.6, 18.0)                      & (89.8, 70.6, 40.0)                      & (91.1, \textcolor{blue}{81.4}, 34.5)                      & (86.8, 33.8, 38.1)                                                                & (\textcolor{blue}{91.9}, 61.9, \textcolor{blue}{45.4})                      & (\textcolor{red}{92.1}, \textcolor{red}{86.8}, \textcolor{red}{48.1})  \\
                             &                                                                                             & VisA                     & (79.6, 56.8, \phantom{1}5.0)           & (95.0, 86.9, 26.3)                      & (\textcolor{blue}{95.5}, \textcolor{blue}{87.0}, 21.3) & (95.1, 71.3, \textcolor{red}{29.2})                                            & (\textcolor{blue}{95.5}, 46.3, 25.0)                                           & (\textcolor{red}{95.6}, \textcolor{red}{90.1}, \textcolor{blue}{28.9}) \\
                             &                                                                                             & BTAD                     & (71.4, 32.8, 11.2)                      & (93.1, 59.8, \textcolor{blue}{46.7}) & (\textcolor{blue}{93.3}, \textcolor{blue}{69.3}, 42.0) & (87.7, 17.1, 36.6)                                                                & (\textcolor{red}{96.1}, 34.4, 46.3)                                            & (\textcolor{blue}{93.3}, \textcolor{red}{75.1}, \textcolor{red}{46.9}) \\
                             &                                                                                             & KSDD2                    & (97.9, 91.2, 17.1)                      & (99.3, 85.4, \textcolor{blue}{58.9})                      & (99.4, \textcolor{blue}{92.7}, 41.8)                      & (96.1, 70.8, 56.4)                                                                & (\textcolor{blue}{99.5}, 68.8, 43.7)                                           & (\textcolor{red}{99.6}, \textcolor{red}{98.4}, \textcolor{red}{72.5})  \\
                             &                                                                                             & DAGM                     & (83.2, 55.4, \phantom{1}3.1)           & (99.0, 83.1, 40.7)                      & (\textcolor{blue}{99.1}, \textcolor{blue}{93.6}, 29.5)                      & (97.0, 40.9, 44.2)                                                                & (96.3, 31.4, \textcolor{blue}{45.9})                                           & (\textcolor{red}{99.5}, \textcolor{red}{98.2}, \textcolor{red}{47.6})  \\
                             &                                                                                             & DTD-Synthetic            & (83.9, 57.8, 11.6)                      & (97.1, 68.7, \textcolor{blue}{62.3})                      & (\textcolor{blue}{97.9}, \textcolor{blue}{92.3}, 52.4) & (94.1, 24.9, 52.8)                                                                & (96.4, 62.8, 62.8)                                                                & (\textcolor{red}{98.2}, \textcolor{red}{94.7}, \textcolor{red}{71.6})  \\ \hline
\multirow{8}{*}{Medical}     & \multirow{3}{*}{\begin{tabular}[c]{@{}c@{}}Image-level\\ (AUROC,F1-Max,\\ AP)\end{tabular}} & HeadCT                   & (83.7, 78.8, 81.6)                      & (93.8, 90.5, 92.2)                      & (95.3, 89.7, 95.2)                                           & (93.3, 86.5, 92.2)                                                                & (\textcolor{blue}{96.8}, \textcolor{blue}{93.4}, \textcolor{blue}{95.3}) & (\textcolor{red}{98.2}, \textcolor{red}{92.9}, \textcolor{red}{98.1})  \\
                             &                                                                                             & BrainMRI                  & (92.0, 84.2, 90.7)                      & (92.8, 88.7, 85.5)                      & (\textcolor{blue}{96.1}, 92.3, 92.3)                      & (94.9, 90.4, \textcolor{blue}{94.2})                                           & (94.1, \textcolor{blue}{93.5}, 93.3)                           & (\textcolor{red}{96.4}, \textcolor{red}{95.5}, \textcolor{red}{97.6})                                         \\
                             &                                                                                             & Br35H                    &(80.5, 74.1, 82.2) & (96.0, 90.8, 95.5) & (\textcolor{blue}{97.3}, \textcolor{blue}{92.4}, \textcolor{blue}{96.1}) & (95.7, 89.1, 95.7) & (96.2, 91.5, 94.2) & (\textcolor{red}{98.1}, \textcolor{red}{95.3}, \textcolor{red}{98.2})                                                              \\ \cline{2-9} 
                             & \multirow{5}{*}{\begin{tabular}[c]{@{}c@{}}Pixel-level\\ (AUROC,PRO,\\ AP)\end{tabular}}    & ISIC                     & (83.3, 55.1, 62.4)                      & (81.6, 29.0, 65.5)                      & (88.4, 78.1, 74.4)                                           & (85.4, \phantom{1}5.3, 70.6)                                                     & (\textcolor{blue}{93.8}, \textcolor{blue}{88.5}, \textcolor{blue}{85.9}) & (\textcolor{red}{94.2}, \textcolor{red}{89.2}, \textcolor{red}{87.0})  \\
                             &                                                                                             & ColonDB                  & (64.8, 28.4, 14.3)                      & (80.3, 58.8, 23.7)                      & (81.9, 71.4, 31.3)                                           & (79.3, \phantom{1}6.5, 26.2)                                                     & (\textcolor{blue}{83.6}, \textcolor{blue}{76.9}, \textcolor{blue}{32.1}) & (\textcolor{red}{84.3}, \textcolor{red}{77.7}, \textcolor{red}{36.3})  \\
                             &                                                                                             & ClinicDB                 & (70.7, 32.5, 19.4)                      & (85.8, 69.7, 39.0)                      & (85.9, 69.6, 42.2)                                           & (84.3, 14.6, 36.0)                                                                & (\textcolor{blue}{89.5}, \textcolor{blue}{79.7}, \textcolor{blue}{56.5}) & (\textcolor{red}{90.1}, \textcolor{red}{80.6}, \textcolor{red}{58.3})  \\
                             &                                                                                             & Endo                     & (68.2, 28.3, 23.8)                      & (85.6, 57.0, 51.7)                      & (86.3, 67.3, 50.4)                                           & (84.0, 10.5, 44.8)                                                                & (\textcolor{blue}{90.2}, \textcolor{blue}{77.0}, \textcolor{blue}{60.7}) & (\textcolor{red}{91.0}, \textcolor{red}{78.9}, \textcolor{red}{65.0})  \\
                             &                                                                                             & Kvasir                   & (69.8, 31.0, 27.5)                      & (82.5, 48.1, 46.2)                      & (81.8, 53.8, 42.5)                                           & (79.4, 12.3, 43.8)                                                                & (\textcolor{blue}{87.6}, \textcolor{blue}{60.9}, \textcolor{blue}{57.2}) & (\textcolor{red}{88.2}, \textcolor{red}{62.2}, \textcolor{red}{62.3})  \\ \bottomrule
\end{tabular}
}
\end{table*}

\section{Experiments}
\subsection{Experimental Setup}
\textbf{Datasets.}\quad We conducted experiments on 14 publicly available datasets, including six industrial datasets: MVTec-AD~\cite{bergmann2019mvtec}, VisA~\cite{zou2022spot}, BTAD~\cite{mishra2021vt}, KSDD2~\cite{bovzivc2021mixed}, DAGM~\cite{wieler2007weakly}, and DTD-Synthetic~\cite{aota2023zero}; and eight medical datasets: HeadCT~\cite{salehi2021multiresolution}, BrainMRI~\cite{kanade2015brain}, BR35H~\cite{zhou2023anomalyclip}, ISIC~\cite{gutman2016skin}, CVC-ColonDB~\cite{tajbakhsh2015automated}, CVC-ClinicDB~\cite{bernal2015wm}, Endo~\cite{hicks2021endotect}, and Kvasir~\cite{jha2019kvasir}. 

\noindent\textbf{Evaluation metrics.}\quad We employ the Area Under the Receiver Operating Characteristic (AUROC), maximum F1 score (F1-max), and Average Precision (AP) for image-level classification; and AUROC, AP, and Per-Region Overlap (PRO) for pixel-level segmentation.

\noindent \textbf{Implementation details.}\quad We employ a pre-trained CLIP (ViT-L-14-336) as the backbone. Patch features are extracted from the 6th, 12th, 18th, and 24th layers. We maintain a zero-shot setting by training on VISA and evaluating on the remaining datasets, and training on MVTec-AD when evaluating VISA. The sequence length $m$ is set to 2, the number of experts $N$ to 8, and the Top-$k$ parameter $k$ to 2. The model is trained for 20 epochs with batch size 32 using Adam (initial learning rate $2 \times 10^{-4}$) on a single NVIDIA A100 GPU.


\subsection{Main Results}

As shown in Table~\ref{table_zsad}, we compare WMoE-CLIP with five state-of-the-art methods: WinCLIP~\cite{jeong2023winclip}, CLIP-AD~\cite{chen2024clip}, AnomalyCLIP~\cite{zhou2023anomalyclip}, AdaCLIP~\cite{cao2024adaclip}, and AA-CLIP~\cite{ma2025aa}. WMoE-CLIP consistently outperforms these baselines. Compared to AA-CLIP, it improves image-level AUROC by 1.9\% and 2.7\% on MVTec-AD and VISA, respectively. Although AA-CLIP adopts a two-stage training strategy, it struggles to capture fine-grained image features. In contrast, our method leverages wavelet transformation to extract frequency information, enabling more discriminative representations. Furthermore, results on eight medical datasets demonstrate WMoE-CLIP’s strong generalization, achieving state-of-the-art performance on both image-level and pixel-level metrics.

Figure~\ref{fig:vis} presents anomaly localization results on industrial and medical datasets. WMoE-CLIP achieves precise localization, particularly in challenging medical scenarios, benefiting from global semantic feature sampling and wavelet-enhanced alignment, which facilitate cross-modal interaction.

\subsection{Ablation Study}
\begin{table}[]
\centering
\caption{Ablation studies with image-level AUROC (I-AUC) and pixel-level AUROC (P-AUC) metrics.}
\label{table_ablation}
\resizebox{0.93\linewidth}{!}{
\begin{tabular}{ccccccc}
\toprule
\multirow{2}{*}{CTDS} & \multirow{2}{*}{WCMA} & \multirow{2}{*}{SA-MoE} & \multicolumn{2}{c}{MVTec-AD} & \multicolumn{2}{c}{VisA} \\ \cline{4-7} 
                      &                       &                         & I-AUC         & P-AUC        & I-AUC       & P-AUC      \\ \hline
                      &                       &                         & 88.6          & 90.8         & 84.8        & 94.3       \\
\ding{51}                    &                       &                         & 89.9          & 91.4         & 85.5        & 94.9       \\
\ding{51}                    & \ding{51}                    &                         & 90.9          & 91.8         & 86.1        & 95.3       \\
\ding{51}                    &                       & \ding{51}                      & 91.5          & 91.6         & 86.7        & 95.1       \\
\ding{51}                    & \ding{51}                    & \ding{51}                      & \textcolor{red}{92.4}          & \textcolor{red}{92.1}         & \textcolor{red}{87.3}        & \textcolor{red}{95.6}       \\ \bottomrule
\end{tabular}
}
\end{table}
We conducted ablation studies on the MVTec-AD and VISA datasets, as shown in Table~\ref{table_ablation}. CTDS models the global distribution, improving image-level and pixel-level AUROC by 1.3\% and 0.6\% on MVTec-AD. WCMA facilitates multimodal interaction and yields improvements of 1.0\% and 0.4\% for image-level and pixel-level AUROC on MVTec-AD. SA-MoE enhances the image-level performance by integrating contextual information, resulting in an improvement of 1.2\% on  VISA. Combining all modules achieves the highest performance, demonstrating the effectiveness of our design.

\section{Conclusion}
In this paper, we propose WMoE-CLIP, a novel CLIP-based wavelet-enhanced mixture-of-experts prompt learning model for ZSAD. To improve the adaptability of text embeddings, CTDS models global features via distribution sampling and integrates rich global information into the prompts. WCMA leverages wavelet decomposition to extract multi-frequency image features. Moreover, SA-MoE dynamically aggregates contextual information, enabling robust perception of global semantics. Extensive experiments on industrial and medical datasets demonstrate the superior performance of our method.

\bibliographystyle{IEEEbib}
\bibliography{strings,refs}

\end{document}